\documentclass[runningheads]{llncs}
\usepackage[T1]{fontenc}
\usepackage{graphicx}
\usepackage{xurl}
\begin{document}
\title{Casting Everything to Online API Services? A Survey of Integrating Localized Speech Recognition Models in Robotic Systems}
\titlerunning{Casting Everything to Online API Services?}
\author{Sheng Li\inst{1}\orcidID{0000-0001-7636-3797} \and
Jing Li\inst{2}\orcidID{0000-0001-8105-5852} \and
Felix Schijve\inst{3} \and
Jun Hu\inst{2}\orcidID{0000-0003-2714-6264} \and
Emilia Barakova\inst{2}\orcidID{0000-0001-5688-4878}}
\authorrunning{S. Li et al.}
\institute{Institute of Science Tokyo, Yokohama, Japan\\
\email{sheng.li@ieee.org}\\ \and
Department of Industrial Design, Eindhoven University of Technology, 5612 AZ Eindhoven, The Netherlands\\
\email{\{j.li2,j.hu,e.i.barakova\}@tue.nl}\\ \and
Department of Biomedical Engineering, Eindhoven University of Technology, 5612 AZ Eindhoven, The Netherlands\\
\email{felixschijve@hotmail.com}}
\maketitle              % typeset the header of the contribution
\begin{abstract}
Automatic speech recognition (ASR) has become a critical component of modern robotic systems because it is one of the most natural and intuitive ways for humans to interact with robots. A commonly used method is to directly use API services online. But is that all we can do? This article provides an overview of how ASR technologies are integrated into various intelligent robots and machines. We discuss the evolution of speech recognition from established approaches to state-of-the-art deep learning models, such as OpenAI's Whisper. We also list large-scale datasets and open source toolkits that have been widely used in both industry and academia. We structure the survey around ASR model families, deployment strategies in robotics (especially ROS-based, cloud-based, and hybrid solutions), and several real-world robotic platforms. Finally, we outline the challenges of deploying robust speech recognition in robots and discuss future directions, including multimodal interaction in diverse and dynamic environments. This paper can help social robotics researchers better navigate the emerging domain of language-based natural human-robot interaction. 

\end{abstract}
\section{Introduction}
Humans naturally communicate through speech; therefore, giving robots the ability to comprehend spoken language greatly improves human-robot interaction. Automatic Speech Recognition (ASR) has been increasingly integrated into contemporary robotic systems, enabling humans to just use speech to control robots. Applications include industrial robots on factory floors and service robots in homes and workplaces. ASR makes technology more approachable and user-friendly by enabling robots to receive commands, respond to inquiries, and deliver information.

The applications of early speech recognition systems were limited due to acoustic models that relied heavily on manually constructed labels and vocabularies. In robotics, speech recognition was further limited by the small computational capacity of the onboard robot hardware and their operation in usually noisy natural environments. In addition, failures in robot design, such as the NAO robot that has its microphones near the fans for cooling, introduced additional difficulties in speech recognition. Finally, the early speech models worked mainly with male voices, and interactions with other user groups, such as children and elderly, for example, were challenging  \cite{barakova2015long,kennedy2017child}. Various dedicated software and hardware solutions have been proposed to alleviate these problems \cite{fukumori2022optical,asaka2024improving} until Deep Learning led to significant advancements in human-robot speech interaction. Under the right circumstances, modern ASR models trained on large audio data sets can attain accuracy comparable to that of humans. For example, with 680,000 hours of training data, OpenAI's Whisper model \cite{radford2023whisper} exhibits strong multilingual transcription capabilities. These sophisticated models can handle a variety of languages and noisy inputs more effectively than previous systems, and they perform on benchmarks that are close to natural human interaction. \cite{kim2024understanding,jiang2024llm,verhelst2025large}.

Alongside algorithmic improvements, the availability of large-scale speech corpora has catalyzed progress. Data sets like LibriSpeech \cite{panayotov2015librispeech}, Mozilla CommonVoice \cite{ardila2020commonvoice}, and Switchboard \cite{godfrey1992switchboard} provide hundreds to thousands of hours of transcribed speech. These resources have been critical for training and evaluating ASR systems, especially for deep neural network approaches. Equally important are open source toolkits such as Kaldi \cite{povey2011kaldi}, ESPnet \cite{watanabe2018espnet}, and SpeechBrain \cite{ravanelli2021speechbrain}, which offer researchers and developers reusable frameworks to build and customize speech recognition models. Such toolkits have lowered the barrier to deploying ASR on robotic platforms by providing pre-built components and recipes for training models on the above datasets.

This paper offers a narrative review of the integration of speech recognition into modern robotics systems. We first review the core ASR technologies and tools. Next, we discuss how speech interfaces are implemented in robot architectures, including both onboard processing and cloud-based solutions. We then list several robotic platforms with speech recognition. Finally, we address the challenges faced when deploying ASR in robotic scenarios and explore future directions.

\section{Method}
As a narrative survey, this paper adopts a pragmatic selection strategy rather than an exhaustive meta-analysis. We prioritize representative works that explicitly describe ASR integration in embodied robotic systems, widely used datasets and toolkits, and platforms that illustrate recurring engineering choices. Accordingly, the literature is organized along three axes: ASR model family, deployment strategy (onboard, cloud-based, or hybrid), and application domain \footnote{There are other dimensions on the systematization: (1) ASR model architectures, (2) large pretrained speech models, (3) datasets, (4) toolkits, (5) integration strategies in robotics, and (6) real-world robot examples. However, in this paper, we focus on three dimensions.}. This lightweight taxonomy helps connect technical choices to robotic constraints and use cases. By weaving these sources together, the survey builds a coherent picture of how speech recognition technologies have developed and how they are applied in robotic systems.

 The strength of the method is that it allows for flexible integration of a wide range of sources and presents a coherent, high‑level picture of a rapidly evolving and multidisciplinary field of ASR for robotics without being constrained by rigid inclusion rules. The limitations of the method are elaborated on in the Discussion section.  

\section{ASR Technologies for Robotics}

\subsection{A Brief Introduction to Speech Recognition for Robots}
Conventional ASR still predominantly adopts hybrid architectures such as Gaussian Mixture Models combined with Hidden Markov Models GMM-HMM \cite{gmmhmm} and Deep Neural
Networks combined with Hidden Markov Models (DNN-HMM) \cite{dnnhmm}, in which the acoustic model, pronunciation lexicon, and language model are tuned separately. End-to-end (E2E) approaches collapse these modules into a single neural network that directly maps variable-length acoustic frames to output symbols (characters, syllables, words, etc.). This streamlined formulation has delivered strong results and has spurred several E2E variants: connectionist temporal classification (CTC) \cite{end2end,eesen}, attention-based encoder–decoder models (e.g., Transformers) \cite{attenearly,lasatten}, lattice-free MMI (LF-MMI) training \cite{e2elfmmi}, and models that jointly optimize CTC and attention objectives (CTC/Attention) \cite{multicmu}.

Transformers have been successfully incorporated into end-to-end ASR systems, yielding competitive performance \cite{transffirst,transfmulti,transfcompare,transfis}. More recently, self- supervised models such as Wav2Vec 2.0 \cite{m3} have become popular because they can be pretrained on large amounts of unlabeled audio, greatly reducing annotation effort. To attain state-of-the-art accuracy, however, these SSL representations must still be fine-tuned on labeled speech using a CTC objective.

Large-scale ASR models pre-trained in vast speech corpora, often called speech foundation models, have recently become prominent in both industry and academia. Training them from scratch is beyond the reach of most researchers due to the enormous data and computational demands, but once released, they greatly streamline downstream development: practitioners can simply fine-tune the pretrained model for their specific domain. Some examples include:

\begin{itemize}
\item 1. OpenAI’s Whisper \cite{openai_whisper} adopts the canonical Transformer encoder–decoder architecture \cite{vaswani2017attention} and is pretrained on roughly 680 k hours of proprietary transcribed audio, enabling state-of-the-art results in automatic speech recognition, speech translation, and language identification.

\item 2. CMU’s OWSM line replicates Whisper-style pretraining entirely with publicly available corpora \cite{owsm1}. The current OWSM v3.1 release keeps the Transformer encoder–decoder framework \cite{owsm2}, whereas the newer OWSM-CTC variant shifts to a Wav2Vec-like encoder optimized for CTC decoding \cite{peng2024owsmctc}.

\item 3. Meta’s Massively Multilingual Speech (MMS) model combines self- supervised pretraining on religious speech corpora with a Wav2Vec-style encoder, achieving ASR coverage for 1,000 + languages while relying on markedly less labeled data than Whisper \cite{mms}.\\

\end{itemize}

Recent studies harness large pretrained language models for generative error correction (GER) in ASR; notably, Zhang et al. \cite{zhang2019investigation} proposed a Transformer-based spelling corrector that reduces substitution errors in Mandarin speech recognition. Building on BERT \cite{devlin2018BERT}, Zhang et al. \cite{zhang2020spelling} used soft masking to link error detection and correction, improving BERT’s ability to spot misspellings. Futami et al. \cite{futami2020distilling} further distilled BERT’s knowledge to create soft labels for acoustic model training. Other studies employ BERT to enhance N-best rescoring \cite{salazar2019masked,shin2019effective}. Beyond ASR, BERT underpins multimodal pretraining—both vision–language \cite{lu2019vilBERT,li2019visualBERT,zhou2020unified,li2020oscar} and speech–language paradigms \cite{baevski2020effectiveness,hsu2021huBERT,wang2020curriculum}. More recently, Chen et al. \cite{Chen2023HyPoradiseAO} integrated large language models directly into the ASR pipeline to further boost recognition accuracy.

Open datasets and collaborative research have accelerated the progress of ASR systems. 

\begin{itemize}
\item In the last decade, hundreds of hours of speech data \footnote{openslr.org} can build solid GMM/HMM, DNN/HMM or early E2E models.
LibriSpeech \cite{panayotov2015librispeech}, for example, contains around 1000 hours of English speech derived from audiobooks and has become a standard benchmark for ASR. CommonVoice \cite{ardila2020commonvoice} extends this concept to dozens of languages through crowd-sourced recordings, providing a massively multilingual corpus. Meanwhile, the Switchboard corpus \cite{godfrey1992switchboard}, with its hundreds of telephone conversations, is especially valuable for developing conversational speech recognition. Training on these datasets allows models to learn rich representations of the spoken language needed for real-world deployment.

\item Recent speech datasets are mostly over tens of thousands of hours of recordings due to the vast parameters of speech foundation models. For English development tasks, Gigaspeech \cite{gigaspeech} has 10,000 hours with high-quality human transcriptions for supervised learning and 33,000+ hours for unsupervised/ semi-supervised learning. Wespeech \cite{wespeech} is a 10,000-hour multi-domain Mandarin corpus for speech recognition. ReasonSpeech \cite{reazonspeech} is another data set of more than 10,000 hours for the Japanese development task.

\item Another feature of recent speech data sets is they are not only limited to speech recognition tasks but can also serve other tasks, such as speech translation, summarization, and QA\footnote{\url{https://iwslt.org/2025/}}.

\end{itemize}

Developers have access to several comprehensive ASR toolkits to experiment with and deploy these advanced models. Kaldi \cite{povey2011kaldi}, one of the first widely adopted speech recognition toolkits, provides a complete suite of algorithms for acoustic modeling, feature extraction, and decoding using finite-state transducers. It has been the backbone of many research systems. More recently, end-to-end focused toolkits such as ESPnet \cite{watanabe2018espnet} and SpeechBrain \cite{ravanelli2021speechbrain} have emerged. ESPnet offers implementations of cutting-edge end-to-end models (for ASR, speech translation, and other tasks) and has recipes that reproduce results on standard corpora. SpeechBrain, built on PyTorch, aims to be an all-in-one toolkit supporting a range of speech technologies within a unified framework. These tools simplify the process of integrating new ASR models into robotic applications by handling the low-level details and providing pre-trained models.

In summary, the technology landscape for speech recognition has matured to the point where accurate, pre-trained models and extensive libraries are readily available. Robotic platforms are an important scenario for these ASR models. The next section discusses how these speech recognition technologies are practically integrated into robotic systems, including commonly used software frameworks and services.

\subsection{Integrating Speech Recognition in Robots}
In a robotic system, speech recognition can be implemented either on the robot itself (onboard processing) or by leveraging external services (offboard processing). Each approach has trade-offs in terms of latency, reliability, privacy, and maintenance.

For onboard implementations, many robots utilise the Robot Operating System (ROS) as middleware. ROS offers packages and nodes for common functionalities, including speech recognition. For example, PocketSphinx \cite{huggins2006pocketsphinx} (a light weight version of CMU Sphinx) has been used in ROS-based robots for offline command-and-control speech recognition. More recently, ROS integrations have been created for Vosk\footnote{\url{https://alphacephei.com/vosk}} (an offline speech API based on Kaldi) and even for OpenAI's Whisper \cite{radford2023whisper}. These ROS packages allow a robot to process microphone input and obtain text transcripts in real time, entirely on-device. The advantage of onboard ASR is that the robot does not depend on network connectivity and can preserve user privacy by not sending audio to external servers. However, running computationally intensive ASR models locally requires sufficient hardware (CPU/GPU) and efficient software optimization.

Alternatively, robots can use cloud-based speech recognition services. Cloud ASR services maintained by major tech companies often achieve very high accuracy by using massive models and continuously learning from large user bases. Examples include Google Cloud's Speech-to-Text API\footnote{\url{https://cloud.google.com/speech-to-text}}, Amazon's Alexa Voice Service (AVS)\footnote{\url{https://developer.amazon.com/alexa}}, and IBM's Watson Speech-to-Text\footnote{\url{https://www.ibm.com/cloud/watson-speech-to-text}}. These services allow robots to stream audio over the Internet and receive transcribed text (and sometimes intent interpretations) in return. Some consumer robots and smart home devices act as front-ends that capture voice commands and rely on cloud ASR for interpretation. For instance, Amazon's own Astro home robot (discussed later) essentially uses Alexa cloud speech recognition as a backbone to understand user requests \cite{seifert2021astro}.

Cloud ASR can offload computation from the robot and leverage cutting-edge models without the need for onboard updates. It can also integrate the robot into a broader ecosystem of voice-enabled services (for example, enabling a robot to answer general questions using cloud knowledge bases). The downsides include the reliance on network connectivity and potentially introducing latency in the interaction. In time-critical or network-denied scenarios, cloud dependency can be problematic. Furthermore, continuous streaming of audio to the cloud raises concerns about data security and user privacy.

A hybrid approach is increasingly being used. It implies performing initial wake-word detection and perhaps basic command recognition on the robot and invoking cloud services for more complex queries. This ensures that simple interactions (like waking the robot by name or executing basic known commands) are fast and offline, while more elaborate tasks (like open-ended questions) rely on the cloud's resources.

To make these trade-offs more explicit, Table~\ref{tab:asr_deployment} summarizes the three dominant deployment patterns discussed in this survey.

\begin{table}[t]
\caption{Structured comparison of the main ASR deployment strategies in robotics.}
\label{tab:asr_deployment}
\centering
\footnotesize
\setlength{\tabcolsep}{3pt}
\renewcommand{\arraystretch}{1.15}
\begin{tabular}{|p{1.45cm}|p{1.45cm}|p{1.55cm}|p{1.75cm}|p{4.35cm}|}
\hline
Strategy & Latency & Privacy & Connectivity & Typical fit in robotics \\
\hline
Onboard & Low and predictable & High & Not required & Offline, privacy-sensitive, or time-critical command and control \\
\hline
Cloud-based & Variable and network-dependent & Lower & Required & Open-domain interaction and access to larger, frequently updated services \\
\hline
Hybrid & Low for routine commands; variable for complex requests & Medium & Partially required & Balance fast local control with cloud support for richer language tasks \\
\hline
\end{tabular}
\end{table}

Table~\ref{tab:asr_deployment} highlights that no single deployment strategy is universally optimal. Onboard ASR is attractive when response time, autonomy, or privacy dominate, whereas cloud ASR is useful when broad coverage and frequent model updates matter most. Hybrid designs are often the most practical compromise in HRI because they preserve responsiveness for common commands while reserving cloud access for linguistically complex requests. Published results remain difficult to compare directly because hardware, noise conditions, and task complexity vary substantially across platforms. For future benchmarking-oriented surveys, studies should ideally report word or character error rate together with end-to-end latency, hardware footprint, connectivity assumptions, and privacy constraints under comparable robotic tasks.

In ROS2, the newer generation of ROS, the speech recognition ecosystem is also evolving. Efforts are underway to incorporate popular ASR models into ROS2 nodes with real-time performance optimizations, making it easier to plug advanced ASR into robotics applications.

In addition to ROS, some robotic platforms have custom software stacks, but still use similar ASR components. Many robotic developers choose to use the aforementioned open-source tools directly (e.g., run a Kaldi or Vosk transcription service locally on the robot) or interface with cloud APIs through HTTP/WS protocols. For example, a mobile robot could use a local Vosk server for offline commands and fall back to Google Cloud Speech-to-Text when online for improved accuracy or extended vocabulary.

In the closely related domain of vehicle systems, there are automotive-specific speech platforms. Companies such as Cerence (a spin-off of Nuance Communications, which is active in robotic voices) specialize in embedded automotive voice assistants that run on the infotainment hardware of cars \cite{schwartz2021cerence}. These systems often use a hybrid ASR approach: an embedded recognizer for driver voice commands (optimized for the car environment and commands like navigation or climate control) paired with cloud connectivity for queries requiring internet information. Similarly, SoundHound Inc. provides the Houndify platform, which has been integrated into some car models for voice control; for example, Houndify powers the voice assistant in certain Honda vehicles \footnote{\url{https://www.businesswire.com/news/home/20200827005197/en/}}\cite{soundhound2020honda}. These automotive ASR solutions illustrate how domain-specific needs (noise robustness, fast response, limited UI) shape the integration of speech recognition.

In summary, integrating ASR into a robot involves choosing between onboard vs. cloud processing, selecting appropriate software frameworks, and handling issues related to real-time performance. Developers have a rich set of ROS modules, APIs, and services to accomplish this integration. The next section will look at concrete examples of robots that use speech recognition as a key interface, highlighting how these technologies come together in practice.

\section{Robotic Platforms with Speech Interaction}
A variety of robotic platforms across different domains rely on speech recognition to interact with users. Below, we highlight several notable examples, ranging from social robots designed to engage with people to industrial and research robots that accept voice commands.

\subsection{Social and Personal Robots}
Humanoid and social robots are perhaps the most intuitive candidates for speech interfaces, as they are often intended to communicate with people in a human-like way. Here is a list of famous examples:\\

\noindent\textbf{Pepper}: One prominent example is Pepper\footnote{\url{https://www.softbank.jp/en/corp/group/sbm/news/press/2015/20150618_01/}}, the humanoid robot introduced by SoftBank Robotics in 2014. Pepper was explicitly designed to interact with humans and is equipped with microphones and speech recognition onboard. It can engage in basic dialogue and respond to questions, making it popular for customer service roles and research on human-robot interaction. SoftBank reports that Pepper was the first personal robot capable of recognizing human emotions from tone of voice and facial cues, highlighting the importance of understanding speech in its design.\\

\noindent\textbf{Misty II}: It is a social robot developed by Misty Robotics (a spin-off of Sphero). Misty II is a tabletop-sized robot with expressive eyes and a voice interface. It is marketed as a developer platform for creating custom robot applications. Out-of-the-box, Misty II supports wake-word detection and command recognition. Developers can program Misty to listen for specific phrases and trigger behaviors. Misty II leverages cloud-based ASR for complex speech tasks but can also run offline keyword spotting. Its creation demonstrates how a personal robot can integrate third-party speech services to achieve flexible interaction \footnote{\url{https://www.theverge.com/2018/5/2/17311124/misty-robotics-sphero-programmable-robot}}.\\

\noindent\textbf{Temi}: It is a wheeled personal assistant robot for home or office use\footnote{\url{https://nocamels.com/2019/01/personal-ai-robot-temi-amazon-alexa/}}. Approximately the size of a small vacuum, Temi can navigate autonomously and has a tablet-like screen for a face. It comes with built-in Alexa integration, effectively giving it powerful speech recognition and natural language understanding through Amazon's voice service. Users can ask Temi questions or give commands as they would to an Alexa smart speaker. By joining Amazon's Alexa Voice Service, Temi offloads its speech processing to the cloud, providing a wide range of voice-enabled features \cite{both2019temi}. This underscores a trend for consumer robots to leverage existing voice assistant ecosystems (Alexa, Google Assistant, etc.) to avoid reinventing speech technology.\\

\noindent\textbf{Amazon Astro}: A recent entrant in the consumer domain is Amazon Astro\footnote{\url{https://www.theverge.com/2021/9/28/22697244/amazon-astro-home-robot-hands-on-features-price}}, a home robot. Astro is essentially an Alexa on wheels — it uses Amazon's speech recognition and voice assistant for all its interactions. Owners can call Astro by saying "Alexa" or "Astro" and then give it commands like checking the security of the home, playing music, or following them with a video call. The speech interface is central to Astro's functionality, as it does not have arms or other complex inputs; voice (and a companion smartphone app) is how users primarily control it. Using the established Alexa infrastructure, Astro demonstrates how a robot can seamlessly integrate into a smart home environment using voice \cite{seifert2021astro}.\\

\noindent\textbf{Anki Vector}: Small companion robots have also experimented with voice input. Anki Vector is a palm-sized robot equipped with a built-in voice assistant. Vector, released in 2018 as the successor to the toy-like Cozmo robot, can respond to greetings, answer trivial questions, and perform simple tasks like taking photos, all triggered by voice commands. It uses onboard wake-word detection ("Hey Vector") and then processes requests via a cloud service. Despite its small size, Vector shows that even very small robots can include always-on microphones and speech processing to improve user engagement \footnote{\url{https://www.theverge.com/2018/8/8/17661902/anki-vector-home-robot-voice-assistant-ai}}\cite{bohn2018vector}.

\subsection{Humanoid and Mobile Robots}
Beyond social robots, speech recognition is playing a role in more general-purpose humanoid robots. \\

\noindent\textbf{Tesla's Optimus} (also known as Tesla Bot) is an upcoming bipedal robot prototype that was announced by Tesla in 2021. While Optimus is still in development. It is intended to eventually perform tasks in human environments, which implies the need for intuitive communication methods. A robot like Optimus is anticipated to employ advanced ASR (potentially based on open-source models such as Whisper) to understand user instructions\footnote{\url{https://spectrum.ieee.org/robotics-experts-tesla-bot-optimus}} \cite{ackerman2022optimus}. Although details are sparse, the inclusion of voice interaction is expected, given Tesla's involvement in AI and its integration of voice control in vehicles. The success of Optimus as a household or factory assistant would likely depend on robust speech recognition to allow untrained users to command it naturally.\\

\noindent\textbf{Boston Dynamics' Spot}: In the realm of legged robots, Boston Dynamics' Spot\footnote{\url{https://www.theverge.com/2019/6/5/18653710/boston-dynamics-first-commercial-robot-spot-demo-amazon-remars-conference-marc-raibert}}(a quadruped robot) provides an interesting case. Spot is primarily controlled via a tablet or joystick, but it can also be programmed to respond to voice commands. Researchers and developers have experimented with mounting microphones and integrating ASR so that Spot can be directed by voice in scenarios where using a handheld controller is inconvenient. For instance, voice commands like "Spot, come here" or "Inspect that area" could trigger autonomous behaviors. As Spot became Boston Dynamics' first commercial product in 2019 \cite{vincent2019spot}, third-party developers have extended its capabilities with software; speech recognition is one such extension for hands-free operation in the field. Especially in applications like search-and-rescue or industrial inspection, voice control could allow an operator to guide Spot while keeping their hands free for other tasks.\\

\noindent\textbf{Autonomous vehicles and service robots in hospitality}: There is another category of robotics, autonomous vehicles and service robots in hospitality, which also incorporates speech interfaces. Many self-driving car prototypes include voice assistants to let passengers request destinations or ask for information during a ride. In such contexts, companies often integrate automotive-grade ASR like Cerence's platform\footnote{\url{https://voicebot.ai/2021/12/22/cerence-co-pilot-introduces-proactive-car-voice-assistant/}} \cite{schwartz2021cerence} to handle voice commands offline in the cabin, supplemented by cloud AI for complex queries. Delivery robots or hotel robots that bring items to guests might use speech to confirm the identity of a person or to allow the person to make requests ("Please send more towels"). These use cases may rely on limited vocabulary speech recognition tuned to specific phrases.\\

In general, these examples illustrate the spectrum of robots that use speech recognition, from small consumer robots to advanced humanoids. The common application is that voice input can make robots more accessible and easier to use, provided that the ASR is accurate and responsive. Each robot leverages speech recognition in a way that suits its purpose: Pepper and Misty for conversational interaction, Temi and Astro by piggybacking on home assistant APIs, Vector to create a bond with the user, Optimus as a future necessity for general use, and Spot or cars for convenient command input.

\section{Challenges and Future Directions}
While speech recognition technology has greatly improved, its integration into robotic systems still faces several challenges. 

\subsection{Acoustic noise and variable environments}
Robots frequently operate in environments with significant background noise (e.g., machinery sounds in a factory, wind outdoors, or multiple people talking in a home). Robustness to noise is thus critical; even state-of-the-art ASR can struggle with distorted audio or overlapping speech. Roboticists are exploring multi-microphone arrays combined with signal processing techniques to perform beamforming and noise cancellation on robots, which can improve ASR input quality.

\subsection{Speaker diversity and language understanding}
Robots may be used by many different individuals, each with unique accents, pitches, and speaking styles. Consequently, the system must generalize well to speakers it has never encountered. Furthermore, if robots are deployed globally, they should ideally understand multiple languages and dialects. This is an area where models like Whisper \cite{radford2023whisper} shine with their multilingual capabilities. However, memory and computational constraints on robots can limit the deployment of larger models. Future advances might focus on developing more efficient models or on-device adaptation techniques that allow robots to learn from their users' speech patterns over time.

\subsection{Latency of system responses}
Humans naturally expect responsive interactions; even brief delays in robotic responses to simple questions can make an interaction feel awkward and unnatural. Ensuring low-latency processing, especially when using cloud services, remain a technical challenge. Techniques like endpointing (deciding when the user has finished speaking) and incremental speech recognition (processing audio as it comes in) can enhance interaction fluidity and responsiveness.

\subsection{Contextual understanding}
Speech recognition produces textual output, but the robot needs to interpret this text within the context of its tasks and environment - a process often referred to as spoken language understanding, or SLU. For example, if a user tells a home robot, "I'm heading out, clean up this mess," the robot needs to translate this command to concrete actions, such as activating a vacuum routine. Achieving this requires linking ASR output to the robot's behavior planning modules. Typically, it involves natural language processing pipelines or machine learning models layered on top of the raw speech recognition. Ensuring that speech commands reliably translate to correct actions (and safely handling misrecognitions) is part of the deployment challenge.

\subsection{Privacy and security}
From a deployment perspective, privacy and security will continue to be important. Users may be uncomfortable with robots that are always listening or sending audio to external servers. Manufacturers will need to be transparent about how speech data is used and possibly offer opt-out features or offline modes to build trust.

\subsection{Toward multimodal interaction}
In practical HRI, speech rarely occurs in isolation. Users also convey intent through gaze, pointing, facial expression, body posture, and the surrounding scene context. As a result, ASR quality should be considered together with multimodal grounding: commands such as ``pick that up'' or ``put it there'' can only be resolved reliably when speech is fused with vision and non-verbal cues. From a survey perspective, this suggests that future comparisons should examine not only transcription accuracy, but also how well speech modules interface with perception, dialogue management, and action execution.

Looking ahead, such multimodal integration will likely be combined with large language models (LLMs) and task planners. A robot can use speech together with vision (e.g., lip reading, object recognition, or gesture interpretation) to improve understanding, while LLM-based dialogue modules can support more natural interaction beyond simple command execution. Researchers are actively exploring end-to-end systems that take speech as input and output actions or responses, effectively merging ASR, NLP, and decision-making into one AI system for the robot.

In conclusion, speech recognition is a cornerstone technology that makes robotics more user-friendly and interactive. Its continued advancement will likely go hand in hand with broader AI improvements. As models become more accurate, compact, and efficient, we will see even greater adoption of speech interfaces in robots of all kinds. Overcoming current challenges will enable robots to understand us reliably in all the noisy and dynamic environments in which we expect them to operate. This will bring us closer to the sci-fi vision of fluent human-robot communication in everyday life.

\section*{Discussion}
In this paper, we provide a narrative survey of how ASR technologies, powered by deep learning and large datasets, are enabling robots to engage in meaningful interactions with humans. Several insights stem from this survey.

First, the robotic speech recognition undergoes a rapid transition from rigid, manual systems to flexible foundation models like Whisper and MMS. In addition, recent open-source efforts are exploring lightweight and extensible speech foundation modeling for robotics, such as the Julius Speech Foundation Model project \footnote{https://github.com/halspeech/julius-speech-foundation-model}, which focuses on modular design and real-time deployment by integrating classical decoding frameworks with modern speech representations.
While early robots were often hindered by limited computational capacity and a failure to recognize diverse voices, such as those of children, modern deep learning has achieved near-human accuracy using massive datasets. However, as these models grow in complexity, researchers must balance between the challenges of hardware demands required for onboard deployment and cloud services. On-device solutions, for instance via ROS packages for Vosk, PocketSphinx, and Whisper, ensure user privacy and reliability in network-denied environments, yet cloud-integrated platforms offer a level of intelligence and scalability that local hardware rarely matches.  Adversely, while cloud services offer immense scale and accuracy, they often suffer from latencies, causing a "turn-taking blind spot" \cite{hri2025}. A central finding in this survey is that no single deployment strategy is universally optimal, and the choice depends on the specific robotic use case. A hybrid approach, performing initial wake-word detection locally while offloading complex queries to the cloud, appears to be the most pragmatic path forward, balancing immediate responsiveness with the depth of global knowledge bases.

Second, despite advancements, deploying robust ASR in robots remains difficult due to acoustic "ego-noise" from robot hardware, environmental background noise, and the diversity of speakers (e.g., accents and age). Furthermore, we can conclude that accurate transcription of the spoken language is only the first step. The major advancement in ASR for robotics is needed in contextual understanding and seamless interaction. This requires ASR to evolve from a standalone ASR module into a central pillar of multimodal HRI. Future systems will likely merge sensory modalities, such as audio with vision, to resolve ambiguities in noisy signals and regulate turn-taking through physical cues like gaze and movement. By merging multiple sensory inputs and reasoning capabilities into a unified AI system, we can move closer to the vision of robots that truly participate in fluent, human-like communication.

The limitations of the study are related to the lack of a reproducible search strategy, since this is not a systematic review. This could have biased the selection of models, datasets, and robotic platforms towards well-known examples, and cannot guarantee comprehensive coverage of the field. In addition, given the fast-moving nature of ASR and robotics, this narrative synthesis can quickly become outdated and offers only limited critical analysis of conflicting findings or gaps in the literature. Given the rapidly evolving nature of ASR in robotics, we believe that a narrative survey is the right choice for the current stage of this research area.

\section*{Conclusion}
In this paper, we reviewed how ASR technologies, powered by deep learning and large datasets, are enabling robots to understand human speech. With frameworks like ROS and cloud APIs, developers can integrate speech interaction into robots more easily than ever before. However, there is still much remaining to make robotic platforms work with state-of-the-art speech recognition and NLP models. A structured comparison of onboard, cloud-based, and hybrid deployment strategies also shows that system design must balance responsiveness, privacy, and computational resources. Real-world and personalized applications, therefore, require careful engineering to handle noise, latency, NLP abilities, and multimodal perception in robotic contexts.

\bibliographystyle{splncs04}
\bibliography{refs_sorted}

\end{document}